\documentclass[aoas]{imsart}
\usepackage{amsmath,amsfonts,amssymb}
\usepackage{mathtools}
\usepackage{graphicx}
\RequirePackage[authoryear]{natbib}
\RequirePackage[colorlinks,citecolor=blue,urlcolor=blue]{hyperref}
\usepackage{multirow}
\usepackage{subcaption}
\usepackage{bm}
\usepackage{bbm}

\begin{document}

\begin{frontmatter}
\title{Using Persistent Homology Topological Features to Characterize Medical Images: Case Studies on Lung and Brain Cancers}
\runtitle{Using Topological Features to Characterize Medical Images}

\begin{aug}
\author[A]{\fnms{Chul} \snm{Moon}\ead[label=e1]{chulm@smu.edu}\orcid{0000-0002-2892-5449}},
\author[B]{\fnms{Qiwei} \snm{Li}\ead[label=e2]{qiwei.li@utdallas.edu}\orcid{0000-0002-1020-3050}}
\and
\author[C]{\fnms{Guanghua} \snm{Xiao}\ead[label=e3]{guanghua.xiao@utsouthwestern.edu}}
\address[A]{Department of Statistical Science, Southern Methodist University \printead{e1}}
\address[B]{Department of Mathematical Sciences, University of Texas at Dallas \printead{e2}}
\address[C]{Quantitative Biomedical Research Center, Department of Population \& Data Sciences and Department of Bioinformatics, University of Texas Southwestern Medical Center \printead{e3}}
\end{aug}

	\begin{abstract}

	  Tumor shape is a key factor that affects tumor growth and metastasis.
	  This paper proposes a topological feature computed by persistent homology to characterize tumor progression from digital pathology and radiology images and examines its effect on the time-to-event data.
	  The proposed topological features are invariant to scale-preserving transformation and can summarize various tumor shape patterns.
	  The topological features are represented in functional space and used as functional predictors in a functional Cox proportional hazards model. The proposed model enables interpretable inference about the association between topological shape features and survival risks.
	  Two case studies are conducted using consecutive 133 lung cancer and 77 brain tumor patients. 
	  The results of both studies show that the topological features predict survival prognosis after adjusting clinical variables, and the predicted high-risk groups have worse survival outcomes than the low-risk groups.
	  Also, the topological shape features found to be positively associated with survival hazards are irregular and heterogeneous shape patterns, which are known to be related to tumor progression. 
	\end{abstract}

	\begin{keyword}
\kwd{Topological Data Analysis}
\kwd{Tumor Shape}
\kwd{Functional Data Analysis}
\kwd{Survival Analysis}
\kwd{Cox Proportional Hazards Model}
    \end{keyword}
\end{frontmatter}
	
	\section{Introduction}
	
	Recent advancements in the field of medical imaging have led to a high-resolution and informative description of human cancer.
	Artificial Intelligence (AI) image processing algorithms, such as deep learning, have been developed to extract information from medical images and have attained comparable achievements with human experts \citep{havaei2017brain,levine2019rise,wang2019pathology}. These AI algorithms also have enabled efficient medical image segmentation, classifying image patches into categories such as tumor and normal regions.
	With such developments, medical images have a significant influence on medical decision-making.
	Radiomics has been developed for decision support by extracting quantitative features of images and providing data for further analyses \citep{gillies2016radiomics,rizzo2018radiomics}.
	The two most common types of radiomic features are texture and shape \citep{bianconi2018}.
	The textural features summarize the area, kurtosis, entropy, and correlation computed from pixel intensity and the gray-level co-occurrence matrix \citep{Haralick1973}.
	The shape features mainly focus on the boundaries of segmented tumor regions and quantify their roughness or irregularities  \citep{Bharath2018,bru2008fractal,kilday1993classifying,bookstein1997morphometric, zhang2020bayesian,crawford2019}.
	However, the existing features provide limited explanations about the tumor shape so that detailed local patterns such as a relationship, distribution, and connectivity between tumor and normal regions are not summarized.
	Also, the features developed to deal with radiographic images could be inadequate in high-resolution pathology images \citep{Madabhushi2016ImageAA,zhang2020bayesian}.
	In this paper, we propose topological features computed by persistent homology to quantify various aspects of tumor shape features in AI-segmented medical images. 
	
	Topological data analysis is a recently emerged area of study that investigates the shape of data using their topological features. 
	Persistent homology is a commonly used topological data analysis tool that analyzes the shape of data with the multi-scale topological lens \citep{Carlsson2009}.
	Persistent homology provides a numeric summary of the shape that is robust to noise and insensitive to metrics \citep{Chazal2017}.
	Persistent homology has been applied to various tumor image analyses, including colorectal tumor region segmentation \citep{Qaiser2016}, clustering of Gleason score of prostate cancer histology \citep{Lawson2019,Berry2020}, and hepatic tumor classification \citep{oyama2019}.
	Only a few studies have focused on survival prediction using topological features of medical images.
	\cite{crawford2019} propose the smooth Euler characteristics to describe the shape of tumor boundaries and develop a functional survival model. 
	Also, \cite{somasundaram2021persistent} show that persistent homology summary features of Computed Tomography (CT) images predict lung cancer patients' survivals. 
	
    We propose topological features computed by persistent homology for analyzing AI-segmented medical images.
    We first develop the distance transform that can reveal the tumor shape of AI-segmented medical images.
    Persistent homology is computed based on the proposed distance transform values.
    Unlike most existing shape features that only focus on a few large segmented tumor regions, the proposed approach quantifies all tumor shape patterns regardless of size.
    The proposed topological features are invariant to the scale-preserving transformation, such as rotation and translation, and are applicable to various types of medical images, including pathology and radiology images. 
    
	Statistical inference on different tumor shape patterns in medical images can be achieved by using the proposed persistent homology features.
	However, persistent homology results are algebraic objects, and this makes it difficult to use them as inputs of machine learning and statistical models.
	Although persistent homology outputs can be summarized as numeric values, they are given as a multiset of intervals, not a vector.
    As a result, several methods have been suggested to vectorize persistent homology results by representing them in different spaces such as Euclidean space \citep{Adams2017}, functional space \citep{Chen2015,Bubenik2015,Adams2017,Berry2020}, and reproducing kernel Hilbert space \citep{Reininghaus2015,kusano2016}.
	In our study, persistent homology results are represented in a functional space to maintain interpretability and assign flexible weights to topological features. Furthermore, the represented functional summary is implemented in a functional survival model as a functional predictor. 

	We develop a Functional Cox Proportional Hazards (FCoxPH) model to characterize the association between functional persistent homology features and survival outcomes. Since \cite{chen2011} proposed the FCoxPH model, various approaches have been developed \citep{Gellar2015,lee2015,qu2016,Kong2018,hao2020}. 
    In our study, the dimension of functional predictors is reduced by Functional Principal Component Analysis (FPCA), one of the key techniques in functional data analysis \citep{yao2005functional}. We select a finite number of basis functions by FPCA and use them in the FCoxPH model.
    We extend the FCoxPH model of \cite{Kong2018} to include multiple functional predictors obtained by persistent homology. The proposed FCoxPH model implements both clinical variables and functional persistent homology features and enables interpretable inference about tumor shape and pattern.
    
    We conduct case studies on lung cancer pathology images and brain tumor Magnetic Resonance Imaging (MRI) images where both image data were collected as part of routine clinical procedures. The results show that the proposed shape features calculated from routine medical images can be used to predict patient prognosis. The predicted high- and low-risk groups show significant differences in survival outcomes for both lung cancer pathology image dataset ($\text{p-value}=4\times10^{-7}$ and $\text{hazard ratio} = 5.381$) and brain tumor MRI image applications ($\text{p-value}=8\times10^{-4}$ and $\text{hazard ratio} = 2.176$). Also, the simulation studies show that the proposed method detects the tumor shape patterns and is robust to false shape information. We find that the irregular tumor shapes and heterogeneous patterns are positively related to risks of death, which coincide with the aggressive tumor patterns. The proposed method enables in-depth shape and pattern analysis on the survival prognosis using topological features of medical images.

	The rest of the paper is organized as follows.
	Section~\ref{sec:method} proposes persistent homology features of AI-segmented medical images, their functional representations, and the FCoxPH model. 
	Section~\ref{sec:simulation} presents a simulation study about false shape information and Section~\ref{sec:case} applies the proposed method to lung cancer pathology and brain tumor MRI image data. 
	Section~\ref{sec:conclusion} concludes the paper and discusses future research topics.

	\section{Topological Shape Analysis for Medical Images}
	\label{sec:method}

	Only a few studies have used topological features of medical images to predict patient survival outcomes.
	First, \cite{crawford2019} use the Smooth Euler Characteristic Transform (SECT) to summarize the shape of the tumor boundary of glioblastoma multiforme (GBM) and predict the survival prognosis using Gaussian Process (GP) regression. 
	Although their approach suggests that tumor shape information paired with its location is useful, it comes with some limitations.
	First, the GP regression model does not consider censored observations because their study is motivated by the dataset without censoring. 
	Second, the SECT is sensitive to the rotation and orientation of images. This makes it difficult to use when medical images do not have pre-defined orientations, such as pathological images.
	Lastly, it is difficult to interpret the GP regression result and provide clinical implications.
	Also, \cite{somasundaram2021persistent} use persistent homology features of the grayscale CT images of lung cancer patients and conduct a Cox Proportional Hazards (CoxPH) model. They summarize persistent homology results using the moments of the topological feature curve. However, the proposed summary loses some information about topological features and lacks interpretability.
	To overcome these challenges, we propose using persistent homology to describe the tumor shape patterns and the FCoxPH model.
	
	\subsection{Persistent Homology Shape Features of Medical Images}

	Persistent homology recently emerged as a powerful analytic tool to characterize shapes \citep{chazal2021introduction}. In persistent homology, the shape of data is quantified using topological features such as connected components, loops, and voids. The connected components and loops are often referred to as the dimension-zero and dimension-one features, respectively. Persistent homology keeps track of the evolution of such topological features in nested shapes defined over a filtration. The persistence of topological features is recorded by their birth and death across the range of the filtration values, and it can be summarized by a persistence diagram, a collection of (birth, death) points in $\mathbb{R}^2$. For a more detailed introduction of persistent homology, see Section~S1 of the Supplementary Material. 
	
\begin{figure}[!ht]
	    \centering
	    \begin{subfigure}{0.22\textwidth}
	        \centering
            \includegraphics[width=0.9\textwidth]{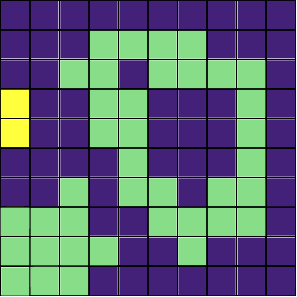}
            \caption{Three-class image}
            \label{subfig:ori.img}
        \end{subfigure}
        \begin{subfigure}{0.22\textwidth}
            \centering
            \includegraphics[width=0.9\textwidth]{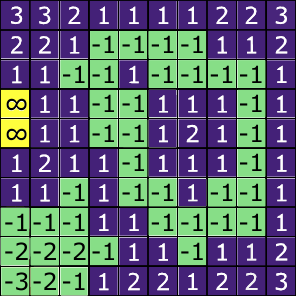}
            \caption{Signed distance}          \label{subfig:sd.img2}
        \end{subfigure}
        \begin{subfigure}{0.22\textwidth}
            \centering
            \includegraphics[width=0.9\textwidth]{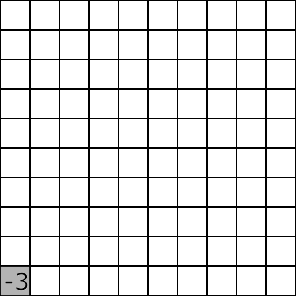}
            \caption{$\mathcal{C}_{-3}$}
            \label{subfig:-3}
        \end{subfigure}
        \begin{subfigure}{0.22\textwidth}
            \centering
            \includegraphics[width=0.9\textwidth]{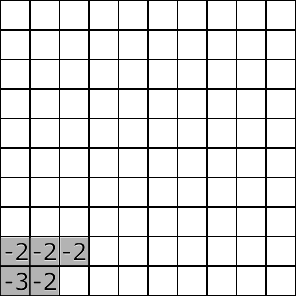}
            \caption{$\mathcal{C}_{-2}$}
        \end{subfigure}
        \begin{subfigure}{0.22\textwidth}
            \centering
            \includegraphics[width=0.9\textwidth]{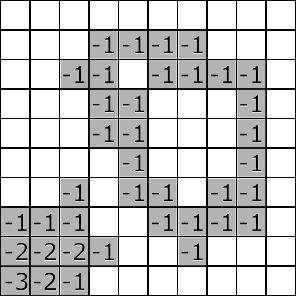}
            \caption{$\mathcal{C}_{-1}$}
            \label{subfig:-1}
        \end{subfigure}
        \begin{subfigure}{0.22\textwidth}
            \centering
            \includegraphics[width=0.9\textwidth]{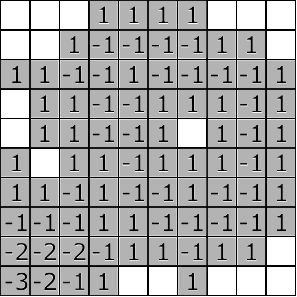}
            \caption{$\mathcal{C}_{1}$}
            \label{subfig:1}
        \end{subfigure}
        \begin{subfigure}{0.22\textwidth}
            \centering
            \includegraphics[width=0.9\textwidth]{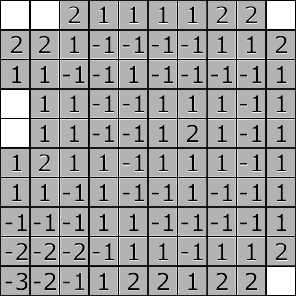}
            \caption{$\mathcal{C}_{2}$}
            \label{subfig:2}
        \end{subfigure}
        \begin{subfigure}{0.22\textwidth}
            \centering
            \includegraphics[width=0.9\textwidth]{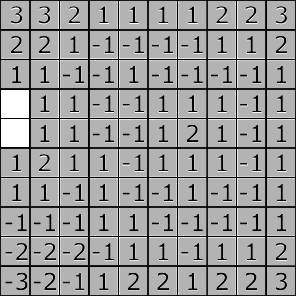}
            \caption{$\mathcal{C}_{3}$}
            \label{subfig:3}
        \end{subfigure}
        \begin{subfigure}{0.4\textwidth}
            \centering
		    \includegraphics[width=0.75\textwidth]{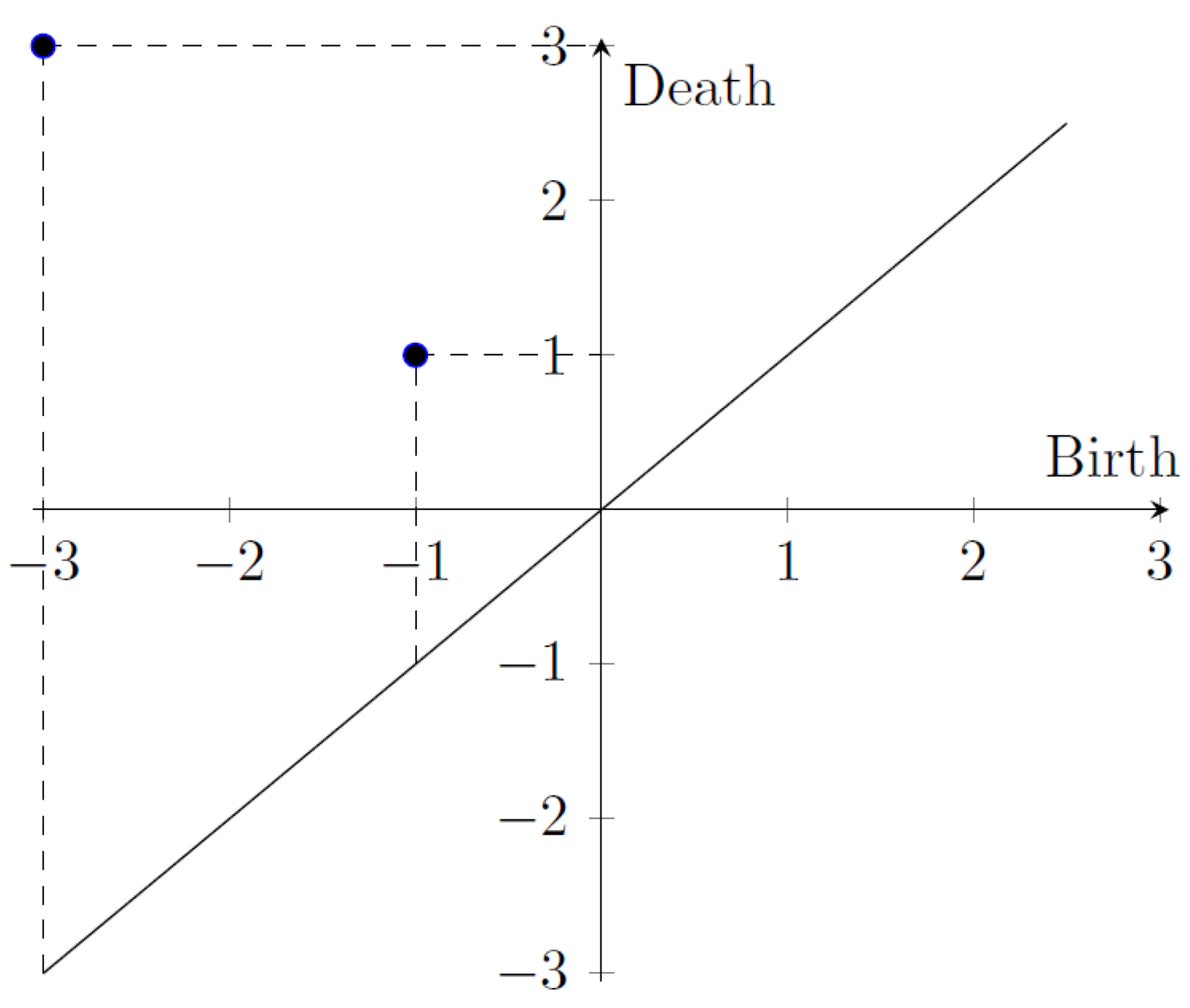}
			\caption{Dimension-zero persistence diagram}
			\label{subfig:pd0}
		\end{subfigure}
		\begin{subfigure}{0.4\textwidth}
		    \centering
		    \includegraphics[width=0.75\textwidth]{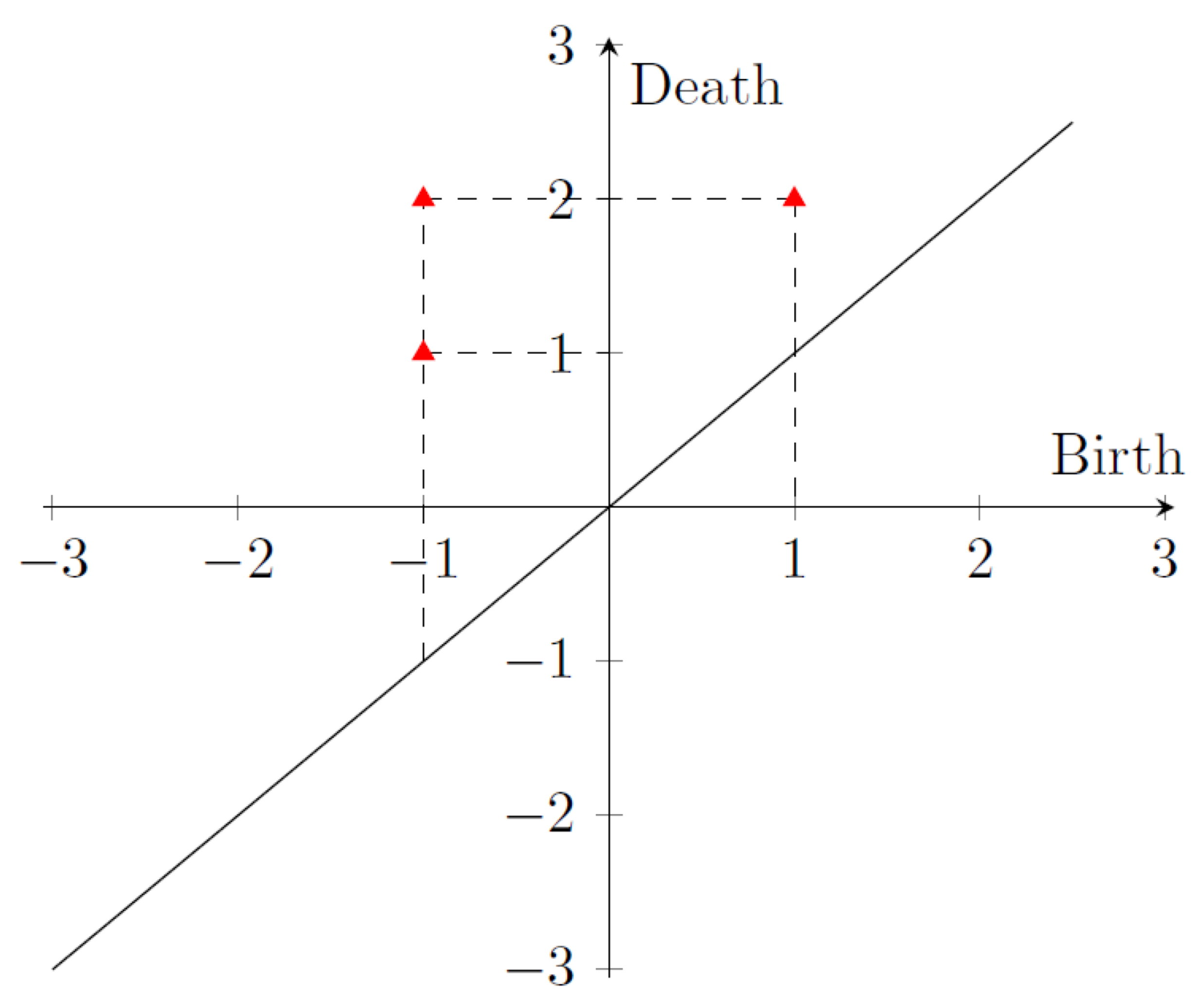}
			\caption{Dimension-one persistence diagram}
			\label{subfig:pd1}
		\end{subfigure}
	    \caption{(a): Three-class example image. The green-, blue-, and yellow-colored pixels are tumor, normal, and empty regions, respectively. (b): Signed distances using the taxicab distance. (c)-(h): A sequence of cubical complices based on the computed signed distance. The pixels included in the cubical complices are marked in gray. (i)-(j): Dimension zero- and one-persistence diagrams.}
	    \label{fig:complex}
	\end{figure}

	Persistent homology is a great tool to summarize the shape of data. However, it is not directly applicable to AI-segmented medical images because the segmented images do not carry natural shape information.
	We propose a Signed Euclidean Distance Transform for the three-class medical images (SEDT-3) to reveal the shape information. The SEDT-3 extends the Signed Euclidean Distance Transform for the binary material images (SEDT-2) of \cite{Robins2016}.
	Suppose that all pixels of medical images are classified into one of the three classes: tumor, normal, and empty regions. 
	For example, Figure~\ref{subfig:ori.img} presents the three-class example image where the green-, blue-, and yellow-colored pixels are tumor, normal, and empty regions, respectively. 
	The three-class image itself does not reveal the shape information, such as connectivity and size information. However, such information can be discovered by applying the distance transform.
	For a given pixel, the SEDT-3 finds the nearest pixel with a different class. Then, it computes the Euclidean distance to that pixel, and assigns the computed distance to the given pixel. The sign of the SEDT-3 value depends on the class of pixel; negative values for tumor region pixels, positive values for normal region pixels, and infinite values for the empty region pixels. 
	The signed distance values have shape information in that they show connectivity and adjacency relationships between neighboring pixels. The SEDT-3 values of Figure~\ref{subfig:ori.img} are computed in Figure~\ref{subfig:sd.img2}. Here, a taxicab metric is used instead of a Euclidean distance for convenience. The number of operations of the SEDT-3 algorithm depends on the image content as well as the size, and the time complexity is up to $O(N^2)$ where $N$ is the number of pixels \citep{fabbri20082d}.
	
	We construct a cubical complex using SEDT-3 to compute persistent homology. 
	The cubical complex is a set of multi-dimensional cubes, such as points, lines, squares, and cubes, that are glued together. 
	The cubical complex allows describing a structure of an image while preserving its topology \citep{Couprie2001}.
	We can obtain multi-scale shape information by computing persistent homology of the sequence of cubical complices.
	Let $\mathcal{C}_{\epsilon}$ be the cubical complex with filtration $\epsilon$. In $\mathcal{C}_{\epsilon}$, the pixels whose assigned distances are less than $\epsilon$ enter the complex. For example, Figures from \ref{subfig:-3} to \ref{subfig:3} present the sequence of cubical complices based on the signed distance in Figure~\ref{subfig:sd.img2} with $\epsilon\in \{ -3, -2, \ldots, 3 \}$. 
	The empty regions are not used to construct cubical complices, so they do not directly affect persistent homology results. 
	For example, the empty region pixels in Figure~\ref{subfig:sd.img2} are not used in $\mathcal{C}_{\epsilon}$ as long as $\epsilon < \infty$. 	

    Persistent homology features are computed using the sequence of cubical complices.
	In Figure~\ref{fig:complex} example, two dimension-zero features (i.e., two connected components) are recorded.
	The first dimension-zero feature appears in $\mathcal{C}_{-3}$ in the bottom-left corner of Figure~\ref{subfig:-3}. The second dimension-zero feature appears in $\mathcal{C}_{-1}$ in the middle of Figure~\ref{subfig:-1} and merges with the first dimension-zero feature in $\mathcal{C}_{1}$. As a result, the first feature appears at $\epsilon=-3$ and persists to the upper bound $\epsilon=3$ and the second feature is born at $\epsilon=-1$ and dies at $\epsilon=1$. 
	Also, we observe three dimension-one features (i.e., three loops) in the example.
	Two holes show up in $\mathcal{C}_{-1}$ in the middle of Figure~\ref{subfig:-1}. The smaller hole disappears in $\mathcal{C}_{1}$ whereas the larger hole is filled in $\mathcal{C}_{2}$. The other dimension-one component appears in $\mathcal{C}_{1}$ and disappears in $\mathcal{C}_{2}$.
    Such birth and death information of topological features is summarized in persistence diagrams in Figures~\ref{subfig:pd0} and \ref{subfig:pd1}. The time complexity of the persistent homology computation algorithm for the cubical complex is $\Theta(3^DN + D2^DN)$, where $D$ is the dimension of the image and $N$ is the number of pixels \citep{wagner2012efficient}.
    
    The proposed approach provides different tumor shape information than existing methods. 
    The previous studies apply persistent homology to the grayscale images and use the intensity level as a filtration \citep{Qaiser2016,Lawson2019,Berry2020,oyama2019,somasundaram2021persistent}. On the other hand, the proposed method applies to the segmented images and uses the signed Euclidean distances as a filtration. 

    Also, the proposed persistent homology features are invariant to scale-preserving transformations, including rotation, translation, and reflection. Our features only depend on the relative locations to the nearest pixels with different classes. Therefore, the proposed approach is automatically invariant to rotations and orientation shifts.
	See Section~S1.2 in the Supplementary Material for more examples.

\subsection{Interpretation of Persistent Homology Computation Result}

\begin{figure}[!ht]
	    \centering
	    \begin{subfigure}{1\textwidth}
	        \centering         \includegraphics[width=0.85\textwidth]{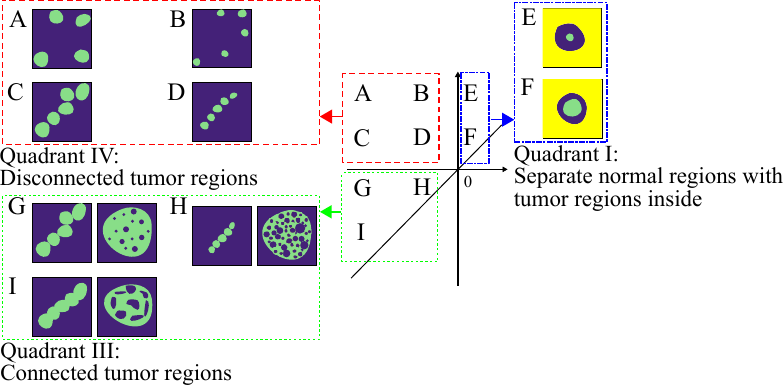}
            \caption{Dimension-zero persistence diagram and the corresponding topological shape feature examples}
            \label{subfig:interpret-dim0}
        \end{subfigure}
        \begin{subfigure}{1\textwidth}
            \centering             \includegraphics[width=0.85\textwidth]{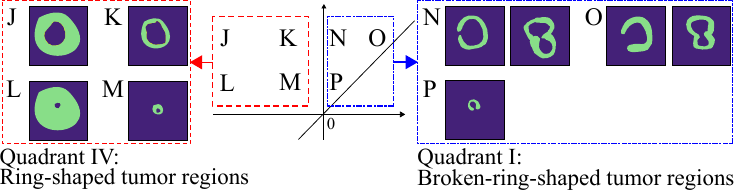}
            \caption{Dimension-one persistence diagram and corresponding topological shape feature examples}
            \label{subfig:interpret-dim1}
        \end{subfigure}
	    \caption{Persistence diagrams of three-class medical images and corresponding tumor shape pattern examples. The green, blue, and yellow colored pixels are tumor, normal, and empty regions, respectively.}
	    \label{fig:features}
	\end{figure}

Various tumor shape patterns can be summarized by the proposed persistent homology approach and they are recorded in different areas of persistence diagrams. Figures~\ref{subfig:interpret-dim0} and \ref{subfig:interpret-dim1} show examples of tumor shape patterns and in which quadrants they are summarized in dimension-zero and dimension-one persistence diagrams, respectively.

    Dimension-zero features appear in the three non-zero quadrants of a dimension-zero persistence diagram: quadrants I, II, and III. 
	First, the scattered tumors are recorded in quadrant II. 
	The larger the tumor regions, the larger birth values are in quadrant II (e.g., A vs. B and C vs. D of Figure~\ref{subfig:interpret-dim0}). Also, larger death values mean that the distances between scattered tumor regions are far from each other (e.g., A vs. C and B vs. D of Figure~\ref{subfig:interpret-dim0}).
	Second, quadrant I represents the separate normal regions that include tumor regions. Death values are proportional to the distance between the boundary of normal regions and the tumor regions inside (e.g., E vs. F of Figure~\ref{subfig:interpret-dim0}).
	Third, the connected tumor regions are summarized in quadrant III. The birth values are the size of tumor regions (e.g., G vs. H of Figure~\ref{subfig:interpret-dim0}), and the death values are the size of the contact area of two tumor regions (e.g., G vs. I of Figure~\ref{subfig:interpret-dim0}). Therefore, the death value indicates how close the connected tumor regions are. 
	The connected tumor regions may have different shapes. For example, they could be a series of overlapped tumor regions (left panels of G, H, and I of Figure~\ref{subfig:interpret-dim0}) or tumor regions that have normal regions inside (right panels of G, H, and I of Figure~\ref{subfig:interpret-dim0}).

	Dimension-one features are summarized in quadrants I and II of a dimension-one persistence diagram. 
	First, the normal regions surrounded by the tumor regions appear in quadrant II. Here, the large negative birth value means the surrounding tumor regions are thick (e.g., J vs. K and L vs. M of Figure~\ref{subfig:interpret-dim1}), and the large death value means the trapped normal regions are large (e.g., J vs. L and K vs. M of Figure~\ref{subfig:interpret-dim1}). 
	Second, the broken-ring-shaped or double-broken-ring-shaped tumor regions are recorded in quadrant I.
	The large birth values imply that there are large gaps in the broken-ring or double-broken-ring-shaped shaped regions (e.g., N vs. O of Figure~\ref{subfig:interpret-dim1}). Also, the large death value means the size of the inside of the broken-ring or double-broken-ring shape is large (e.g., N vs. P of Figure~\ref{subfig:interpret-dim1}).

	The size of persistent homology features is measured by the radius of the largest circle that can be placed inside. Also, dimension-zero and dimension-one features are not exclusive; the pixels used to construct dimension-zero features can be used to build dimension-one features and vice versa. For example, the ring-shaped and broken-ring-shaped tumor regions in Figure~\ref{subfig:interpret-dim1} are also counted as dimension-zero features.  
	
	We note that the persistence diagram is not an invertible function of the shape; we can summarize the shape using the persistence diagram but cannot recover the shape from the persistence diagram. However, one may identify the location of the features on the image using the persistence diagram \citep{obayashi2018persistence}.

    \subsection{Functional Representation of Persistent Homology Shape Features}
    
    Although persistence diagrams include topological persistence signal information, it is not easy to use them directly as input in data analysis.
	This is because persistence diagrams are not in a common data type that most statistical models use.
    In our study, we represent persistence diagrams in a functional space inspired by \cite{Chen2015} and \cite{Adams2017}.
    
    Let $P=\{(b,d)\in \mathbb{R}^2: b<d \}$ be a persistence diagram such that $(b,d)$ denotes the birth and death values. The persistence surface function $\rho_P$ of the persistence diagram $P$ can be defined as 
    $
        \rho_{P} (x,y) = \sum\limits_{(b,d) \in P} g_{(b,d)}(x,y) \cdot w(b,d),
    $
    where $x$ and $y$ are the $(x,y)$-coordinates of the persistence surface function, $g_{(b,d)}$ is a smoothing function for $(b,d) \in P$, and $w(b,d)\geq 0$ is a non-negative weight function.
    A persistence surface function is a stable representation of a persistence diagram; persistence surface functions are robust to small perturbations of $(b,d)$ points in persistence diagrams \citep{Adams2017}. Figure~S3 in the Supplementary Material illustrates the persistence diagram example and its functional representation.
    
    Two persistence surface functions will be obtained when the proposed persistent homology approach is applied to 2D medical images: dimension-zero and dimension-one. We denote persistence surface functions of dimension-zero and dimension-one persistence diagrams of image $i$ by $X_i^0$ and $X_i^1$, respectively.
    
    The weight function $w(b,d)$ allows assigning different importance to $(b,d)$ points in persistence diagrams.
    Various weights have been proposed, including the weights that depend on the persistence of features \citep{Chen2015,Adams2017,kusano2016} and birth, death, and persistence values \citep{Berry2020}.
    In our study, we use three weights: 1) maximum distance weight $w_m(b,d)=\max\{|b|,|d|,d-b\}$, 2) linear weight $w_l(b,d)=d-b$, and 3) arctangent weight $w_a(b,d)=\arctan(\mathcal{C}(d-b)^\mathcal{D})$ with $\mathcal{C}=\mathcal{D}=1$. The maximum distance weight assigns heavier weights to features far from the origin of persistence diagrams. On the other hand, the linear and arctangent weights are proportional to persistence. We only present the results using the maximum distance weight for the rest of the paper. The results using linear and arctangent weights are included in Section~S3.1 of the Supplementary Material. 
    The results of the three weights are generally similar, but they are not the same because each weight emphasizes different aspects of topological features. For example, the linear and arctangent weights focus on the long-persisting features over the short-living features, whereas the maximum distance weight can highlight a few short-surviving features far from the origin.

    We use the Gaussian smoothing function $g_{(b,d)}(x,y)=\frac{\exp\left[-\left((x-b)^2+(y-d)^2\right)\right]}{\sigma^2}$ in the persistence surface function. The Gaussian smoothing function requires the selection of smoothing parameter $\sigma$. We denote the smoothing parameters for dimension-zero and dimension-one persistence surface functions as $\sigma_0$ and $\sigma_1$.
    In our study, the smoothing parameters are determined by the Leave-One-Out Cross-Validation (LOOCV) using 625 combinations of $\sigma_0$ and $\sigma_1$ over the 2D grid in $[0.2, 5]$ with a step size 0.2. The LOOCV results of case studies are given in Section~S3.2 in the Supplementary Material.

    \subsection{Cox Regression Model for Functional Data}
    \label{subsec:fcoxph}
    
    The CoxPH model \citep{Cox1972} is a commonly used model to investigate the association between patients' survival prognosis and predictor variables. 
    The hazard function with a $p$-dimensional scalar predictor $Z=(z_{1},\ldots,z_{p})^T$ has the form 
        $h(t)=h_0(t)\exp\left( Z^T \gamma \right)$,
    where $h_0$ is the baseline hazard function and $t \in [0, \tau ]$ for $0<\tau<\infty$. 
    We aim to conduct the FCoxPH model that uses a set of clinical predictors $\gamma$ and two functional predictors $X^0$ and $X^1$ as
    \begin{equation}
        h(t)=h_0(t)\exp\left( Z^T \gamma + \int X^0(u)\alpha(u) du + \int X^1(v)\beta(v) dv \right).
        \label{eq:hazard}
    \end{equation}
    
    The objective of the FCoxPH model is to determine the unknown coefficients $\gamma$, $\alpha$, and $\beta$. 
    Due to the infinite dimensionality of functional data, dimension reduction is often required. 
    We use FPCA to represent functional data 
    in a lower-dimensional space.
    Let $X_i^0$ and $X_i^1$ be the persistence surface functions of medical image $i$ and $\mu^0(u)=E[X_i^0(u)]$ and $\mu^1(v)=E[X_i^1(v)]$ be the mean functions of $X_i^0$ and $X_i^1$. By the spectral decomposition, the covariance functions can be represented as  $\text{Cov}\left(X_i^0(u),X_i^0(u')\right)=\sum\limits_{j=1}^\infty  \lambda_{j}\phi_j(u)\phi_j(u')$ and $\text{Cov}\left(X_i^1(v),X_i^1(v')\right)=\sum\limits_{k=1}^\infty \delta_{k}\pi_k(v)\pi_k(v')$, where $\{ \lambda_j \}_{j\geq 1}$ and $\{\delta_k\}_{k\geq 1}$ are non-increasing eigenvalues and $\{\phi_j\}_{j\geq 1}$ and $\{\pi_k\}_{k\geq 1}$ are orthonormal eigenfunctions of $X_i^0(u)$ and $X_i^1(v)$.
    By the Karhunen-Lo\`{e}ve expansion \citep{karhunen1947lineare,loeve1946fonctions}, the persistence surface functions can be expressed as
    $X_i^0(u) = \mu^0(u) + \sum\limits_{j=1}^\infty  \xi_{ij}\phi_j(u)$ and
    $X_i^1(v) = \mu^1(v) + \sum\limits_{k=1}^\infty  \zeta_{ik}\pi_k(v)$,
    where $\xi_{ij}=\int \left( X_i^0(u)-\mu^0(u) \right)\phi_j(u)du$ and $\zeta_{ik}=\int \left( X_i^1(v)-\mu^1(v) \right)\pi_k(v)du$ are the Functional Principal Component (FPC) scores of dimension-zero and dimension-one, respectively. The FPC scores have mean zero $E[\xi_{ij}]=E[\zeta_{ik}]=0$ with covariances  $E[\xi_{ij}\xi_{ij'}]=\lambda_j \mathbbm{1}(j=j')$ and $E[\zeta_{ik}\zeta_{ik'}]=\delta_k \mathbbm{1}(k=k')$, where $\mathbbm{1}(\cdot)$ is an indicator function.
    We can approximate functional data $X_i^0 \approx \mu^0(u) + \sum\limits_{j=1}^q  \xi_{ij}\phi_j(u)$ and $X_i^1 \approx \mu^1(v) + \sum\limits_{k=1}^r \zeta_{ik}\pi_k(v)$, where $q$ and $r$ are the selected number of eigenfunctions. Then the FCoxPH model (\ref{eq:hazard}) can be approximated as
    \begin{equation}
        h_i(t) \approx h_0^*(t)\exp\left( Z_i^T \gamma + \sum\limits_{j=1}^{q} \xi_{ij}\alpha_j +
        \sum\limits_{k=1}^{r} \zeta_{ik}\beta_k \right),
        \label{eq:model}
    \end{equation}
    where $h^*_0(t)=h_0(t)\exp\left( \int \mu^0(u)\alpha(u)du + \int \mu^1(v)\beta(v)dv \right)$. 
    The dimension of predictors is reduced to $p+q+r$ in (\ref{eq:model}). We can obtain the estimated coefficients using the selected eigenfunctions and their estimated coefficients as $\hat{\alpha}(u) \approx \sum\limits_{j=1}^q \hat{\alpha}_j\hat{\phi}_j(u)$ and $\hat{\beta}(v) \approx \sum\limits_{k=1}^r \hat{\beta}_k\hat{\pi}_k(v)$. 
    
    The FCoxPH model that incorporates the functional tumor shape predictors enables interpretable inference about the association between shape patterns and survival outcomes. The estimated functional coefficients $\hat{\alpha}(u)$ and $\hat{\beta}(v)$ inform which parts of persistence surface functions are associated with survival risks.
    The findings also can be presented by figures that plot the estimated coefficients on the space of a persistence surface function. By comparing the estimated coefficients plots and the topological features summarized in Figure~\ref{fig:features}, one can see which types of shape patterns may contribute the most to hazard prediction.

    Two criteria are used to select the number of FPCs for the FCoxPH models. 
    First, the percentage of variance explained by the FPCs is used to determine $q$ and $r$ for validation tests. 
    Let $PV^0(q)=\sum\limits_{j=1}^{q} \lambda_j /\sum\limits_{j=1}^{\infty}\lambda_j$ and $PV^1(r)=\sum\limits_{k=1}^{r} \delta_k /\sum\limits_{k=1}^{\infty}\delta_k$ be the percentages of variances explained by $q$ dimension-zero and and $r$ dimension-one FPCs, respectively. For a given variability threshold $C$, the minimum number of FPCs that exceed the threshold: $q=\min \{q: PV^0(q)>C\} $ and $r=\min \{r: PV^1(r)>C\}$. 
    Second, the Akaike information criterion (AIC) is used to choose the number of the FPCs for estimating the FCoxPH models \citep{yao2005functional}. 
    Let $L(\gamma_1, \ldots, \gamma_p, \alpha_1, \ldots, \alpha_q, \beta_1,\ldots,\beta_r \mid q, r)$ denote the partial likelihood function of the FCoxPH model (\ref{eq:model}) given the number of the FPCs $q$ and $r$. 
    The AIC value of the FCoxPH model is $AIC(q,r) = 2(q+r) - 2\log\{L(\hat{\gamma}_1, \ldots, \hat{\gamma}_p, \hat{\alpha}_1, \ldots, \hat{\alpha}_q, \hat{\beta}_1,\ldots,\hat{\beta}_r \mid q, r)\}$.
    We determine an optimal number of components $q$ and $r$ by computing the AIC under the various combinations of $q$ and $r$.
    For the tied event times, we use the approximation method of \cite{Efron1977} to adjust the partial likelihood.
    \cite{Kong2018} show by simulation studies that the percentage of variance criterion and the AIC are suitable for validation tests and model estimation for the FCoxPH model, respectively.

    \section{Simulation Study}
    \label{sec:simulation}
    
\begin{figure}[!ht]
        \centering
        \begin{subfigure}{0.45\textwidth}
        \begin{subfigure}{0.28\textwidth}
            \includegraphics[width=1\textwidth]{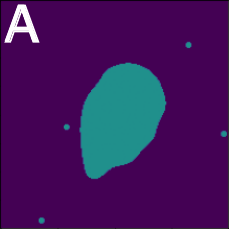}
            \includegraphics[width=1\textwidth]{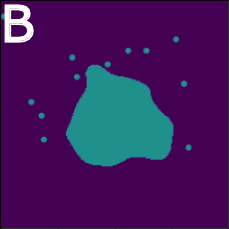}
        \end{subfigure}
        \begin{subfigure}{0.56\textwidth}
            \includegraphics[width=1\textwidth]{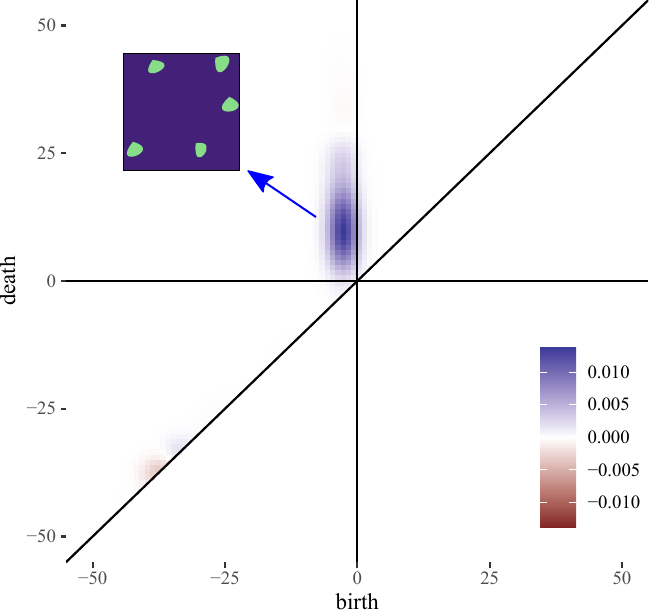}
        \end{subfigure}
        \caption{Scenario 1}
        \label{subfig:sim1}
        \end{subfigure}
        \centering
        \begin{subfigure}{0.45\textwidth}
        \begin{subfigure}{0.28\textwidth}
            \includegraphics[width=1\textwidth]{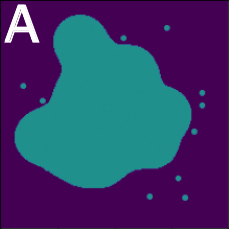}
            \includegraphics[width=1\textwidth]{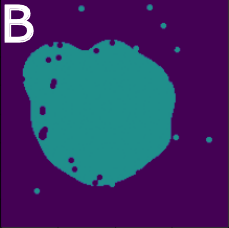}
        \end{subfigure}
        \begin{subfigure}{0.56\textwidth}
            \includegraphics[width=1\textwidth]{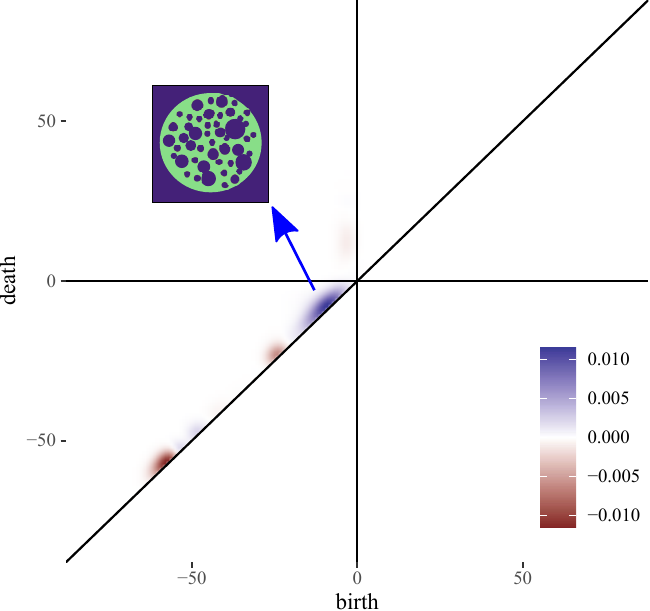}
        \end{subfigure}
        \caption{Scenario 2}
        \label{subfig:sim2}
        \end{subfigure}
        \caption{The tumor image examples (left) and the average estimated dimension-zero coefficient functions of the functional Cox proportional hazards model (right) for each scenario. In the tumor image examples, the green and purple colored pixels are tumor and normal regions. The estimated dimension-zero coefficient functions are plotted on the space of the dimension-zero persistence surface function. The examples tumor shape patterns that correspond to the non-zero coefficients are presented.}\label{fig:simfig}
    \end{figure}

    We conduct simulation studies under two scenarios. For each scenario, we randomly generate 100 binary tumor images of size $200\times200$ pixels from two groups of A and B, respectively, and create 30 independent datasets. 
    In scenario 1, the main tumor region is created by applying Gaussian smoothing and thresholding to 50 points sampled from the bivariate normal distribution. Then, up to 5 and 20 random points are added as small tumor regions to groups A and B, respectively.
    The two groups mainly differ by the number of small disconnected tumor regions, summarized in quadrant IV of the dimension-zero persistence diagram, as illustrated in Figure~\ref{subfig:interpret-dim0}.
    In scenario 2, larger main tumor regions are created similar to scenario 1, and up to 20 small tumor regions are added to both groups. For group B, up to 50 holes are created around the boundary of the main tumor region. The tumor region with holes corresponds to topological features in quadrant III of the dimension-zero persistence diagram, as shown in Figure~\ref{subfig:interpret-dim0}. The simulated tumor image examples of two groups are given in the left panels in Figures~\ref{subfig:sim1} and \ref{subfig:sim2}.

     We transform the simulated images by the SEDT for two-class (SEDT-2). The SEDT-2 is a simpler version of the SEDT-3; it does not assign infinite values because the generated images do not have empty regions. Interpretation of the topological shape features of binary images is similar to interpretation of the three-class images shown in Figure~\ref{fig:features}, except that there are no topological features that summarize separate non-tumor regions (e.g., E and F in Figure~\ref{subfig:interpret-dim0}). 
     
     Persistent homology is computed as illustrated in Section~\ref{sec:method}, and the results are represented as persistence surface functions using smoothing parameters $\sigma_0=\sigma_1=2$. We replace the dimension-zero persistence result with infinite death value $(b,\infty)$ with $(b,b)$.

     The survival times are generated from the CoxPH model by using the inverse probability method of \cite{bender2005generating}. We assume the hazard function $h(t)=0.01  \exp\left(0.8 \times \mathbbm{1}(\text{B})\right)$, where $\mathbbm{1}(\text{B})$ is the indicator of group B. As a result, higher hazard is assigned to group B. We also generate two clinical variables, age and sex, under the same conditions for both groups: age is sampled from the Poisson distribution with mean 40, and sex is randomly determined with the probability of 0.5. 
     
     We conduct two models, the CoxPH and FCoxPH models. The CoxPH model only uses two clinical variables, whereas the FCoxPH model uses the functional persistent homology basis selected by the AIC in addition to two clinical variables. We examine the validity of topological features in the FCoxPH model for predicting the survival outcome. We conduct the chi-square test with $q+r$ degrees of freedom that uses the $q+r$ number of coefficients of functional basis,
	\begin{eqnarray*}
	    H_0:&&\alpha_1=\alpha_2=\ldots=\alpha_q=\beta_1=\ldots=\beta_r=0 \nonumber \\
	    H_1:&&\alpha_j\neq 0 \text{ or } \beta_k\neq0 \text{ for at least one } j\in\{1,2,\ldots q\} \text{ or } k\in\{1,2,\ldots r\}.
	    \label{eq:hypothesis}
	\end{eqnarray*}
	The percentage of variance explained by functional basis is used to select $q$ and $r$, and the variance threshold of $C=90\%$ is used.
    
    The simulation study results show that the proposed topological features reflect the differences of tumor patterns between the two groups. 
    The average p-values of the chi-square tests are small, 0.016 and 0.026 for scenarios 1 and 2, respectively. Also, the average p-value of the Wald test of the FCoxPH models is 0.004 for both scenarios, which is smaller than those of the CoxPH models, 0.557 and 0.475 for scenarios 1 and 2, respectively. The distribution of the p-values is summarized in Figure~S5 in the Supplementary Material.
    
    We also conduct simulation studies to examine false positives using images from a single group, either group A or B. The results imply that the proposed model robustly detects the false shape features. See Section~S2 in the Supplementary Material for a more detailed summary.
    
    The functional coefficients of the proposed model show which shape patterns contribute the most to hazard. The average estimated dimension-zero functional coefficients are shown in the right panels in Figures~\ref{subfig:sim1} and \ref{subfig:sim2}.
    The coefficients show how topological features are associated with the survival prognosis; the blue-colored and red-colored areas are positively and negatively related to the hazard function, respectively.
	Thus, if a larger number of topological features appear in the blue (red) region, it is associated with higher (lower) hazard. 
    For scenario 1, the topological features summarized in quadrant IV of the dimension-zero persistence diagram in Figure~\ref{subfig:sim1} are positively associated with hazard. This is consistent with group B having a larger number of small disconnected tumors than group A. Also, for scenario 2, the topological features that appear in quadrant III near the origin of the dimension-zero persistence diagram in Figure~\ref{subfig:sim2} are related to higher hazard. This corresponds to the tumor regions created by small holes. The red-colored region around $(-30,30)$ in Figure~\ref{subfig:sim2} shows the difference of the size of tumor regions. Because the holes are added to the main tumor region in group B, its size, measured by the radius of the largest circle that can be placed inside, is smaller than that of group A.
	The coefficient of dimension-one functional predictors is not reported because they are not selected for more than half of the datasets.
	
    \section{Case Studies}
    \label{sec:case}
    \subsection{Application to Lung Adenocarcinoma Pathology Images}
    \label{subsec:lung}
\begin{figure}[!ht]
	    \centering
	    \begin{subfigure}{0.28\textwidth}
	        \centering
            \includegraphics[height=0.70\textwidth]{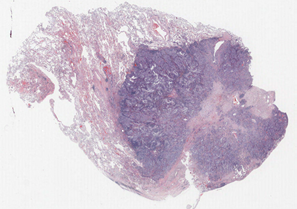}
            \caption{Pathology image}
        \end{subfigure}
	    \begin{subfigure}{0.28\textwidth}
	        \centering
            \includegraphics[height=0.70\textwidth]{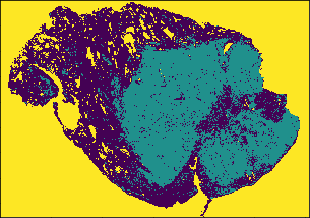}
            \caption{Three-class image}
        \end{subfigure}
        \begin{subfigure}{0.28\textwidth}
            \centering
            \includegraphics[height=0.70\textwidth]{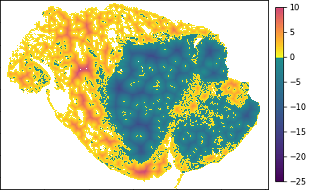}
            \caption{Signed distance}
        \end{subfigure}
	    \caption{The lung adenocarcinoma pathology image (left), three-class pathology image segmented by \cite{Wang2018} (middle), and signed Euclidean distance transformed image (right).
	    In the three-class image, the green, purple, and yellow colored pixels are tumor, normal, and empty regions. In the signed distance transformed image, tumor regions are colored from blue to green and normal regions are colored yellow to red.}
	    \label{fig:pathology}
	\end{figure}

    Lung cancer is one of the most deadly cancers \citep{Siegel2020}. One of the most common types of lung cancer is adenocarcinoma which accounts for about 40\% of all lung cancers \citep{zappa2016} and has various morphological features \citep{Matsuda2015}.
	We use 230 pathology images of 133 lung adenocarcinoma patients in the National Lung Screening Trial (NLST) data. All images are obtained under 40X magnification, and the median size of images is 24,244$\times$19,261 pixels.
	The image patches of size 300$\times$300 pixels (75$\times$75 microns) are segmented into three classes of the tumor, normal, and empty regions using a deep convolutional neural network (CNN) \citep{Wang2018}.
    For example, for a given NLST pathology image of size 30,000$\times$30,000, the CNN model generates the segmented image of size 100$\times$100.
	We implement an additional pre-processing step to remove noise: a single pixel is considered to be noise when its class is different from the surrounding eight singular-class pixels, and the noise pixels are reclassified to the class of the surrounding pixels.
	The denoised three-class images are transformed using the SEDT-3. Figure~\ref{fig:pathology} presents the pathology image, the segmented three-class image of \cite{Wang2018}, and the SEDT-3 image. 
	
	A sequence of cubical complices is constructed using the SEDT-3 values as filtration, and persistent homology is computed using GUDHI \citep{gudhi}. We replace the dimension-zero result with infinite death value $(b,\infty)$ with $(b,b)$ and exclude the dimension-one result with infinite death value $(b,\infty)$ to remove features related to empty regions.
	The median number of the computed features per image slice is 694 for dimension-zero and 1,768 for dimension-one. The ranges of persistence diagrams are $(-41, 11)$ for dimension-zero and $(-19, 26)$ for dimension-one.
	The smoothing parameters $\sigma_0=1.8$ and $\sigma_1=0.4$ are used for the persistence surface functions.
	
	\subsubsection{Model Estimation}

	We construct the CoxPH model only using the following scalar clinical predictors: age, sex, smoking status, cancer stage (I to IV), cancer grade (0 to 4), and tumor size.
	The size of the tumor is measured by the number of tumor pixels in an image slice. Also, we fit the FCoxPH model using the functional predictors and the same clinical predictors used in the CoxPH model. 
	For both models, we assume the baseline hazard function $h_0(t)=1$. 
	The AIC is used to choose the number of FPCs for the FCoxPH model and yields four FPCs: one FPC for dimension-zero and three FPCs for dimension-one.
	We note that the SECT of \cite{crawford2019} is not used because it is sensitive to rotation and translation and is not suitable for pathology image application.

\begin{table}[!ht]
	\caption{The outputs of the Cox proportional hazards (CoxPH) and functional Cox proportional hazards (FCoxPH) models of the lung cancer adenocarcinoma patients. The `PH' is the abbreviation of persistent homology.}
	\centering
\begin{tabular}{l|rr|rr}
\hline
                                 & \multicolumn{2}{c|}{CoxPH}                         & \multicolumn{2}{c}{FCoxPHl}                        \\ \cline{2-5} 
                                 & \multicolumn{1}{c}{coef.} & \multicolumn{1}{c|}{p-value} & \multicolumn{1}{c}{coef.} & \multicolumn{1}{c}{p-value} \\ \hline
Age                              & 0.070                     & 0.036                        & 0.062                     & 0.082                       \\
Smoker vs. non-smoker            & 0.014                     & 0.967                        & -0.181                    & 0.597                       \\
Male vs. female                  & -0.032                    & 0.927                        & -0.020                    & 0.958                       \\
Tumor size                       & $<$0.001                  & 0.026                        & $<$0.001                  & 0.796                       \\
Stage II vs. stage I             & -0.327                    & 0.596                        & 0.136                     & 0.836                       \\
Stage III vs. stage I            & 1.042                     & 0.015                        & 1.066                     & 0.014                       \\
Stage IV vs. stage I             & 1.519                     & 0.003                        & 1.786                     & 0.001                       \\
Grade 1 vs. grade 0              & -1.557                    & 0.065                        & -1.969                    & 0.039                       \\
Grade 2 vs. grade 0              & -0.752                    & 0.308                        & -0.912                    & 0.285                       \\
Grade 3 vs. grade 0              & -0.611                    & 0.394                        & -0.625                    & 0.447                       \\
Grade 4 vs. grade 0              & -16.980                   & $<$0.001                     & -17.150                   & $<$0.001                    \\
PH dimension 0, 1$^{\text{st}}$ FPC & -                         & -                            & 0.015                     & $<$0.001                    \\
PH dimension 1, 1$^{\text{st}}$ FPC & -                         & -                            & 0.001                     & 0.002                       \\
PH dimension 1, 2$^{\text{nd}}$ FPC & -                         & -                            & -0.002                    & 0.001                       \\
PH dimension 1, 3$^{\text{rd}}$ FPC & -                         & -                            & -0.001                    & 0.047                       \\ \hline
\end{tabular}
\label{tb:model.res}
\end{table}
	
	Table~\ref{tb:model.res} shows the results of the CoxPH and FCoxPH models. 
	The clinical variables behave similarly for both models except for the tumor size. The p-value of the tumor size variable increases in the FCoxPH model as the functional predictors are added, suggesting that one of the functional predictors may include the tumor size information.
	The p-values of the Wald tests are close to zero for both models.

	\begin{figure}[!ht]
	    \centering
	    \begin{subfigure}{0.49\textwidth}
	        \centering
            \includegraphics[width=0.65\textwidth]{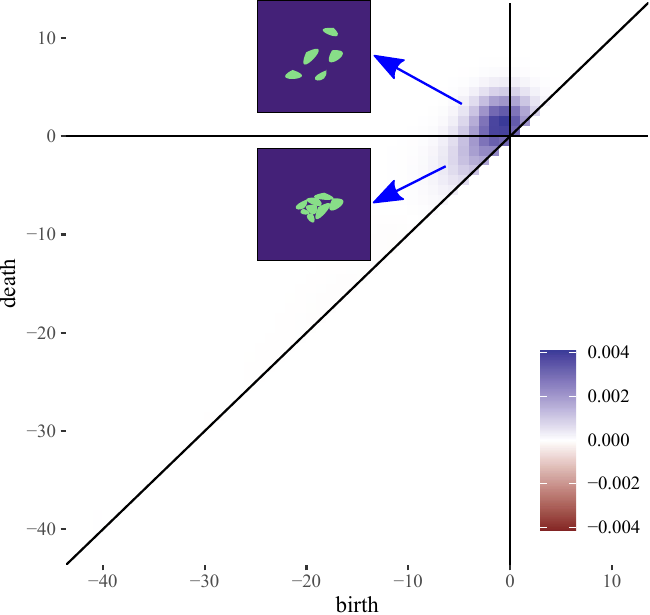}
             \caption{Dimension-zero functional coefficient $\hat{\alpha}(u)$}
            \label{subfig:dim0coef}
        \end{subfigure}
        \begin{subfigure}{0.49\textwidth}
            \centering
            \includegraphics[width=0.65\textwidth]{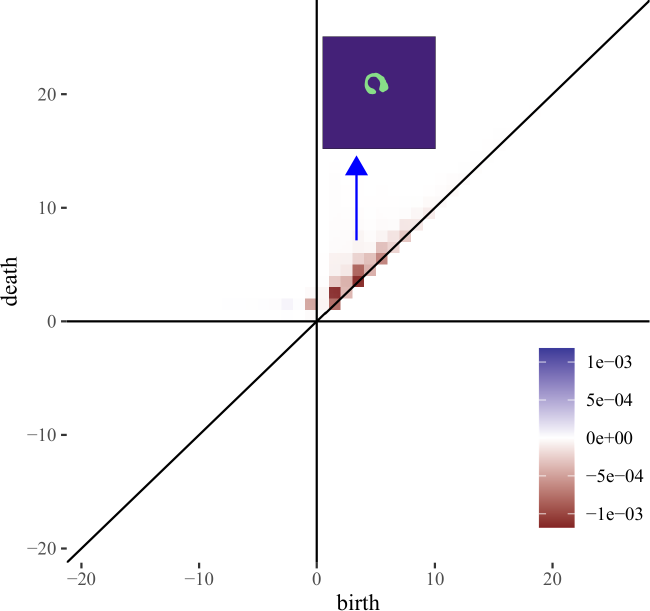}
            \caption{Dimension-one functional coefficient $\hat{\beta}(v)$}
            \label{subfig:dim1coef}
        \end{subfigure}
	    \caption{Estimated coefficient functions $\hat{\alpha}(u)$ and $\hat{\beta}(v)$ of the functional Cox proportional hazards model of the lung adenocarcinoma patients. The estimated coefficient functions are plotted on the spaces of the dimension-zero and dimension-one persistence surface functions. The examples tumor shape patterns that correspond to the non-zero coefficients are drawn.}
	    \label{fig:coef}
	\end{figure}

	Figure~\ref{fig:coef} shows the estimated functional coefficients. 
    The dimension-zero topological features summarized in the blue-colored areas in quadrants II and III of Figure~\ref{subfig:dim0coef} correspond to the aggressive tumor patterns.
	First, the colored area in quadrant II in Figure~\ref{subfig:dim0coef} represents the small-sized scattered tumor regions close to each other (e.g., D of Figure~\ref{subfig:interpret-dim0}).
	Also, quadrant III of Figure~\ref{subfig:dim0coef} represents the relatively small connected tumor shapes. 
	The small connected tumor features are spotted where multiple tumor regions spread inside normal regions (e.g., H of Figure~\ref{subfig:interpret-dim0}). 
	These patterns imply a fast spread of tumors.
	
	On the other hand, the shapes that appear in the red-colored regions in quadrant I of Figure~\ref{subfig:dim1coef} match with less aggressive tumor patterns. 
	The red-colored region corresponds to broken-ring-shaped tumor regions (i.e., see the examples of areas N, P, and O in Figure~\ref{subfig:interpret-dim1}). 
	These shapes require an ample-sized normal region that could be surrounded by the tumor but not invaded by the tumor.
	
	These dimension-one features do not appear when small tumor regions penetrate the normal region inside the broken-ring-shaped tumor regions. 
	Therefore, the broken-ring-shaped tumor may not likely appear where the small tumor regions are densely populated.
	These results coincide with the findings that tumor shape complexity and heterogeneous spread are associated with prognosis \citep{yokoyama1991,miller2003,Chatzistamou2010,vogl2013,grove2015} and tissue transport properties \citep{soltani2012,sefidgar2014}.

	Another interesting observation is that the relatively small-sized topological features, approximately a radius of up to $10$ pixels ($750$ microns), have a strong association with the hazard function. 
	We see that the estimated non-zero functional coefficients, the colored regions in Figure~\ref{fig:coef}, are close to the origin compared to the ranges of persistence surface functions. The topological features near the origin correspond to small-sized features that have small birth and death values. 
	This indicates that the valuable information for our lung cancer study obtained by the proposed persistent homology approach relates mainly to local shapes and patterns.

	\subsubsection{Validity Test}

	We conduct the chi-square test that uses the $q+r$ number of coefficients of functional basis. The variance threshold of $C=90\%$ is used and two dimension-zero and two dimension-one functional basis are selected ($q=r=2$).
	The computed p-value of the chi-square test of degrees of freedom four is $7\times10^{-15}$, suggesting that the topological features are a strong signal. 
	Therefore, aside from the clinical variables included in the model, the topological features offer additional information in predicting lung cancer patients' survival outcomes.

	\subsubsection{Prediction using Cross-validation}
	We predict the risk scores using the LOOCV for the CoxPH and FCoxPH models. For a given pathology image $I \in \{1,2,\ldots,230 \}$, the models are trained for the rest 229 images. The risk score of $I$ is predicted using the trained model. We repeat it for all 230 images, and the predicted risk scores are averaged for each patient. We assign 133 patients into two groups of 66 high-risk patients and 67 low-risk patients using the median patient-wise risk score.

		\begin{figure}[!ht]
	    \centering
	    \begin{subfigure}{0.35\textwidth}
	        \centering \includegraphics[width=1\textwidth]{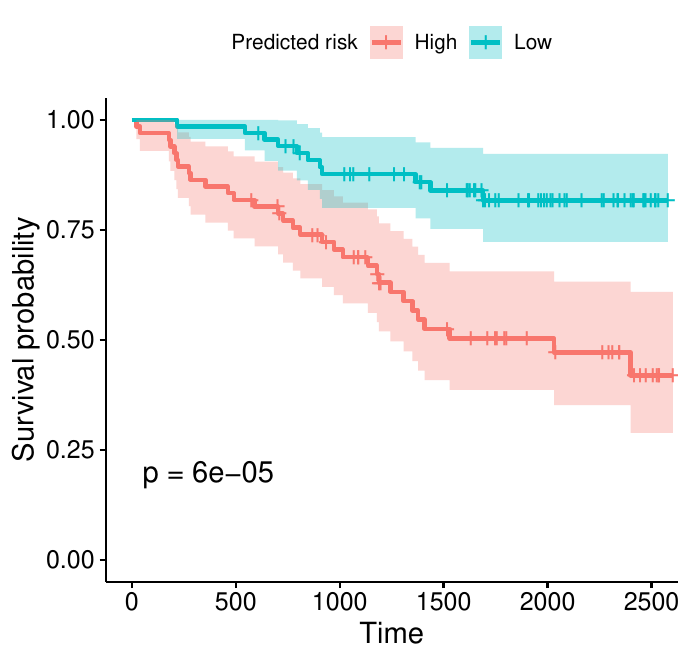}
            \caption{CoxPH model}
        \end{subfigure}
        \begin{subfigure}{0.35\textwidth}
            \centering
            \includegraphics[width=1\textwidth]{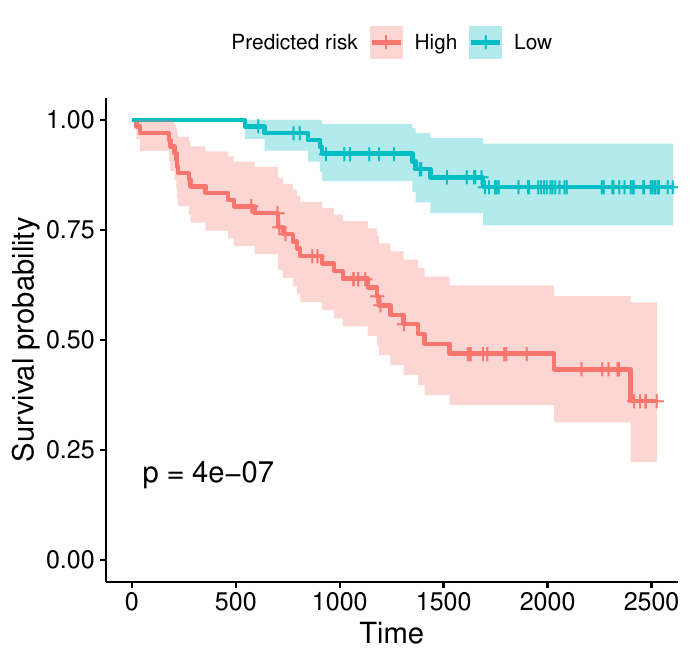}
            \caption{FCoxPH model}
        \end{subfigure}
	    \caption{The Kaplan–Meier plots for the high- and low-risk groups of the Cox proportional hazards (CoxPH) and functional Cox proportional hazards (FCoxPH) models of the lung adenocarcinoma patients.}
	    \label{fig:KMplot}
	\end{figure}
	
	The proposed FCoxPH model shows a better separation between the two groups over the CoxPH model. The Kaplan-Meier plots of the high- and low-risk groups predicted by the two models are presented in Figure~\ref{fig:KMplot}. The p-value of the log-rank test of the FCoxPH model is $4\times10^{-7}$, which is smaller than that of the CoxPH model $6\times10^{-5}$. The hazard ratios between the predicted high-risk and low-risk groups of the FCoxPH and CoxPH models are $5.381$ and $3.682$, respectively. 

	\subsubsection{Prediction under False Shape Information}
	\label{subsec:falsepositive}
	
	A simulation study is conducted to examine whether our method detects false shape information. We create new images by randomly arranging the tumor and normal pixels from the existing three-class images. The rearranged image retains the same proportions of the tumor, normal, and empty regions as the original image but loses original shape information. Figure~S10 in the Supplementary Material shows the three-class pathology image and rearranged image.
	We generate a total of 100 datasets, each with 230 rearranged images. Persistent homology is computed, and computation outputs are represented as a persistence surface function using the same setting. For each dataset, we predict the risk scores using the LOOCV, assign high- and low-risk groups, and compute the p-values of the log-rank tests. 
	
	The prediction results imply that the proposed FCoxPH model detects false shape information. The false shape functional predictors are not selected in the FCoxPH model by the AIC among 94.43\% of the 23,000 models. Also, most p-values of the log-rank tests of the FCoxPH models with rearranged images are similar to the p-value of the CoxPH model; 77 out of 100 p-values are the same as that of the CoxPH model ($6\times10^{-5}$), and only six out of 100 p-values differ from the CoxPH model's p-value by more than $4\times10^{-4}$. Figure~S11 in the Supplementary Material shows the distribution of the p-values of the log-rank tests of the FCoxPH models with rearranged images. This suggests that the false shape information has a limited impact on the FCoxPH model.

	\subsection{Application to GBM MRI Images}
    \label{sec:5}
    
\begin{figure}[!ht]
\centering
\centering
	    \begin{subfigure}{0.22\textwidth}
	        \centering
             \includegraphics[height=\textwidth]{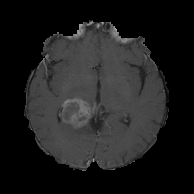}
            \caption{MRI image}
            \label{subfig:brain.mri}
        \end{subfigure}
	    \begin{subfigure}{0.22\textwidth}
	        \centering
            \includegraphics[height=\textwidth]{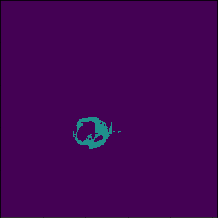}
            \caption{Binary image}
            \label{subfig:brain.binary}
        \end{subfigure}
        \begin{subfigure}{0.22\textwidth}
        \centering
          \includegraphics[height=\textwidth]{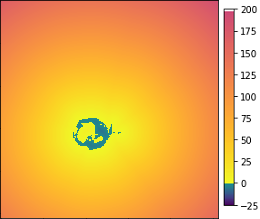}
            \caption{Signed distance}
            \label{subfig:brain.sd}
        \end{subfigure}
 \caption{The T1-weighted contrast-enhanced MRI image of GBM patient (left), segmented binary image (middle), and signed distance transformed image (right). In the binary image, the green and purple pixels are the tumor and non-tumor regions, respectively. In the signed distance transformed image, the tumor regions are colored from blue to green and the normal regions are colored yellow to red.}
 \label{fig:gbm}
\end{figure}

    GBM is the most common malignant grade IV brain tumor  \citep{surawicz1999descriptive}.
    GBM is distinguished from lower-grade astrocytomas (grades II and III) by its accelerated growth rate. The rapid outward growth of GBM develops necrosis, which is considered a hallmark of GBM. On T1-weighted contrast-enhanced imaging MRI, most GBM cases show a ring-shaped enhancement made of hypervascular tissues with a necrotic region at the center \citep{zhu2000quantification,henson2005mri}. 
    Figure~\ref{subfig:brain.mri} shows the MRI image of the GBM patient with a ring-enhancing mass.
    The presence of necrosis is a significant prognosis factor \citep{nelson1983necrosis}, and clinical studies show that the degree of necrosis is negatively associated with a survival rate \citep{hammoud1996prognostic,raza2002necrosis}. 
    However, due to its irregular shape, multifocal enhancement, and the existence of multiple small lesions, evaluation of GBM shapes using MRI images could be difficult
    \citep{eisenhauer2009new,upadhyay2011conventional}.
    
    We use 77 GBM patients' MRI scans obtained from The Cancer Imaging Archive (TCIA) \citep{scarpace2016radiology} and their clinical data retrieved from The Cancer Genome Atlas (TCGA) \citep{cancer2008comprehensive}. The MRI images are segmented into two classes of the tumor and non-tumor regions using the Medical Imaging Interaction Toolkit with augmented tools for segmentation \citep{chen2017fast}. Each patient has approximately 23 to 25 segmented MRI images and the spaces between the MRI images are not the same.
    The size of images is either $256 \times 256$ or $512 \times 512$. 
    We only use 1,190 MRI images that have more than 100 tumor pixels because some images do not include tumor regions that are not large enough.
    The segmented binary images are transformed by the SEDT-2. We note that 2D MRI images are used because the vertical spaces between the MRI images are not the same. However, our method can easily be extended to 3D images.
    Figures~\ref{subfig:brain.binary} and \ref{subfig:brain.sd} show the segmented binary image and distance transformed by the SEDT-2.

    The topological shape features in the MRI images are obtained by computing persistent homology of the cubical complex based on the SEDT-2 values. Especially the ring-shaped enhancements of the GBM patients' MRI images are recorded as dimension-one topological shape features. As illustrated in Figure~\ref{subfig:interpret-dim1}, the broken-ring- and unbroken-ring-shaped masses appear in quadrant I and II of the dimension-one persistence diagram, respectively. 
    
    For the GBM images, the number of topological shape features is much smaller than that of the lung cancer pathology images.
    The median of the number of computed persistent homology features per image slice is 16 for dimension-zero and 17 for dimension-one, respectively.
    This is because MRI images are scanned in lower resolution than pathology images, and GBM tumors are smaller and simpler than lung cancer regions.
    Also, the persistent homology features obtained from the images of size 512$\times$512 are divided by two for a consistent comparison with the images of size 256$\times$256. 
    The ranges of persistent homology shape summaries are $(-22, 26)$ for dimension-zero and $(-6, 32)$ for dimension-one.
    We represent the topological shape features as persistence surface functions using the smoothing parameters $\sigma_0=1.6$ and $\sigma_1=0.4$.  
    The mean persistence surface function is used to represent each patient's tumor shape information.

	\subsubsection{Model Estimation}

We fit the CoxPH and FCoxPH models to predict the overall survival of the GBM patients. We also compare the proposed topological shape features with the SECT of \cite{crawford2019}. Although the GP model is proposed to predict survival outcomes using the SECT in \cite{crawford2019}, we implement it as a functional predictor in the FCoxPH model (FCoxPH-SECT) for a direct comparison. Let $X^{\text{SECT}}$ be the SECT functional predictor. Then, the FCoxPH-SECT model becomes
$h(t)=h_0(t)\exp\left( Z^T \gamma + \int X^{\text{SECT}}(u)\omega(u) du \right)$.
By implementing the SECT to the FCoxPH model, its coefficients can also be interpreted using the invertibility of the Euler characteristic \citep{wang2021statistical}. However, a complete interpretation of the tumor shape pattern is not possible because the mean SECT is used to represent shape information. 
For all models, four clinical predictors are used: age, gender, Karnofsky performance score (KPS), and tumor size. The size of the tumor is calculated by the median of the number of tumor pixels in each patient's images. In the FCoxPH model, four dimension-one FPCs are selected by AIC. 

   \begin{table}[!ht]
    \caption{The outputs of the Cox proportional hazards (CoxPH) model and functional Cox proportional hazards models using the smooth Euler characteristic transform (FCoxPH-SECT) and persistent homology topological feature (FCoxPH) of glioblastoma multiforme patients. The `PH' is the abbreviation of persistent homology.}
\begin{tabular}{l|rr|rr|rr}
\hline
                                    & \multicolumn{2}{c|}{CoxPH model}                         & \multicolumn{2}{l|}{FCoxPH-SECT model}                   & \multicolumn{2}{c}{FCoxPH model}                        \\ \cline{2-7} 
                                    & \multicolumn{1}{c}{coef.} & \multicolumn{1}{c|}{p-value} & \multicolumn{1}{l}{coef.} & \multicolumn{1}{l|}{p-value} & \multicolumn{1}{c}{coef.} & \multicolumn{1}{c}{p-value} \\ \hline
Age                                 & 0.037                     & 0.002                        & 0.048                     & $<$0.001                     & 0.040                     & 0.002                       \\
Male vs. female                     & -0.204                    & 0.465                        & -0.659                    & 0.044                        & -0.270                    & 0.328                       \\
Karnofsky performance score         & -0.019                    & 0.034                        & -0.028                    & 0.013                        & -0.012                    & 0.240                       \\
Tumor size                          & $<$0.001                  & 0.178                        & $<$0.001                  & 0.078                        & $<$0.001                  & 0.985                       \\
SECT, 1$^{\text{st}}$ FPC           & -                         & -                            & 0.001                     & 0.301                        & -                         & -                           \\
SECT, 2$^{\text{nd}}$ FPC           & -                         & -                            & -0.010                    & 0.320                        & -                         & -                           \\
SECT, 3$^{\text{rd}}$ FPC           & -                         & -                            & -0.033                    & 0.022                        & -                         & -                           \\
SECT, 4$^{\text{th}}$ FPC           & -                         & -                            & 0.018                     & 0.194                        & -                         & -                           \\
SECT, 5$^{\text{th}}$ FPC           & -                         & -                            & 0.013                     & 0.310                        & -                         & -                           \\
SECT, 6$^{\text{th}}$ FPC           & -                         & -                            & 0.033                     & 0.032                        & -                         & -                           \\
SECT, 7$^{\text{th}}$ FPC           & -                         & -                            & 0.038                     & 0.013                        & -                         & -                           \\
PH dimension 1, 1$^{\text{st}}$ FPC & -                         & -                            & -                         & -                            & 0.037                     & 0.050                       \\
PH dimension 1, 2$^{\text{nd}}$ FPC & -                         & -                            & -                         & -                            & -0.023                    & 0.310                       \\
PH dimension 1, 3$^{\text{rd}}$ FPC & -                         & -                            & -                         & -                            & -0.012                    & 0.659                       \\
PH dimension 1, 4$^{\text{th}}$ FPC & -                         & -                            & -                         & -                            & 0.082                     & 0.014                       \\ \hline
\end{tabular}
\label{tb:model.res2}
\end{table}

Table~\ref{tb:model.res2} summarizes the results of the CoxPH, FCoxPH-SECT, and FCoxPH models. Age has small p-values in all models. The p-values of the Wald tests are $0.001$, $3\times10^{-4}$, and $7\times10^{-4}$ for the CoxPH, FCoxPH-SECT, and FCoxPH models, respectively.

\begin{figure}[!ht]
	    \centering
        \begin{subfigure}{0.49\textwidth}
            \centering
            \includegraphics[width=0.7\textwidth]{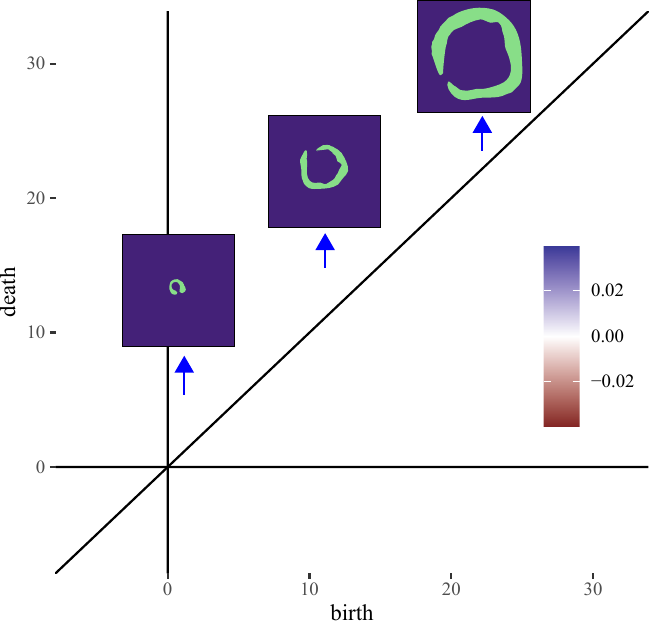}
        \end{subfigure}
	    \caption{Estimated coefficient function $\hat{\beta}(v)$ of the functional Cox proportional hazards model of the glioblastoma multiforme patients. The estimated coefficient function is plotted on the space of the dimension-one persistence surface function. The examples tumor shape patterns that correspond to the non-zero coefficients are presented.}
	    \label{fig:coef.brain}
	\end{figure}

Figure~\ref{fig:coef.brain} presents the estimated coefficient of dimension-one persistence surface function $\hat{\beta}$ of the GBM patients. The blue-colored areas indicate that a larger number of small- and large-sized broken-ring-shaped features are associated with higher risks. These patterns are most likely to correspond to the large necrotic center and heterogeneous enhancement outside of the rim.

\subsubsection{Validity Test}
We conduct the chi-square test to examine the validity of functional topological shape features obtained from the brain tumor MRI images. We use the variability threshold $C=90\%$ and three dimension-zero FPCs and 10 dimension-one FPCs are selected. The p-value of the chi-square test with degrees of freedom of 13 is $1\times10^{-6}$. This implies that the proposed topological shape features are informative for predicting the survival outcomes of the GBM patients after adjusting the clinical variables.

\subsubsection{Prediction using Cross-validation}
We compare the prediction results of three models: CoxPH, FCoxPH-SECT, and FCoxPH. We obtain the predicted risk scores by using the LOOCV and designate high- and low-risk groups of 38 and 39 patients, respectively.

	\begin{figure}[!ht]
	    \centering
	    \begin{subfigure}{0.32\textwidth}
            \centering
            \includegraphics[width=\textwidth]{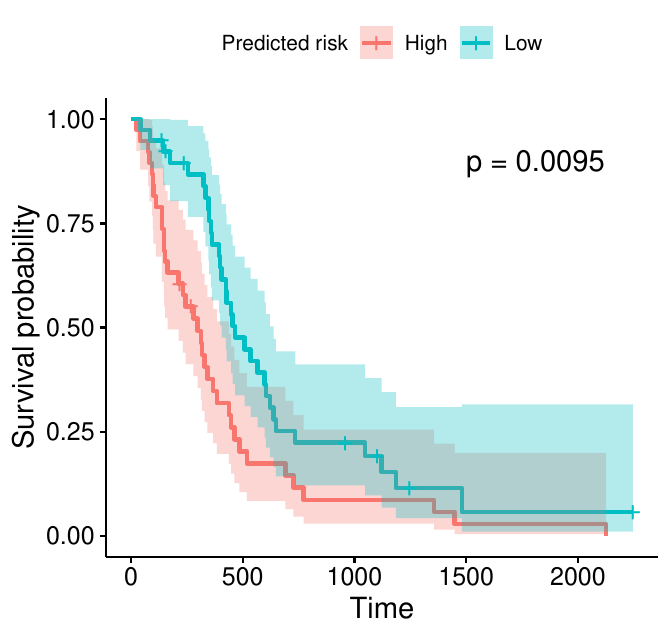}
            \caption{CoxPH model}
        \end{subfigure}
	    \begin{subfigure}{0.32\textwidth}
	        \centering
            \includegraphics[width=\textwidth]{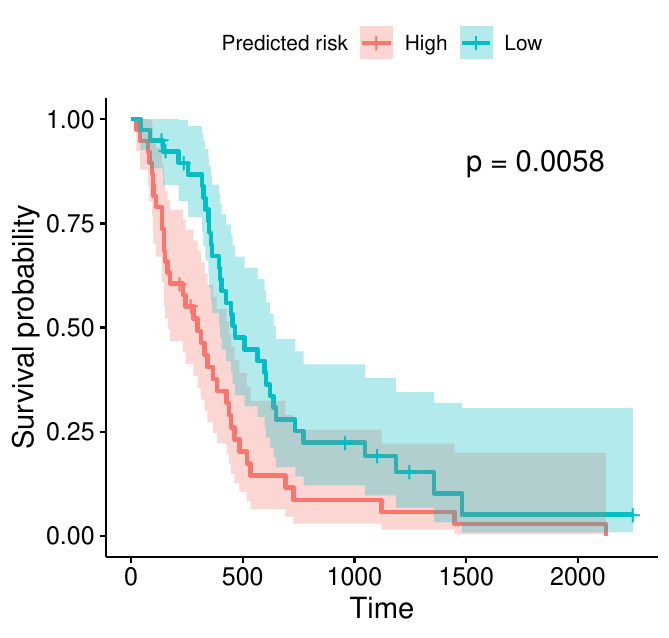}
            \caption{FCoxPH-SECT model}
        \end{subfigure}
        \begin{subfigure}{0.32\textwidth}
            \centering
            \includegraphics[width=\textwidth]{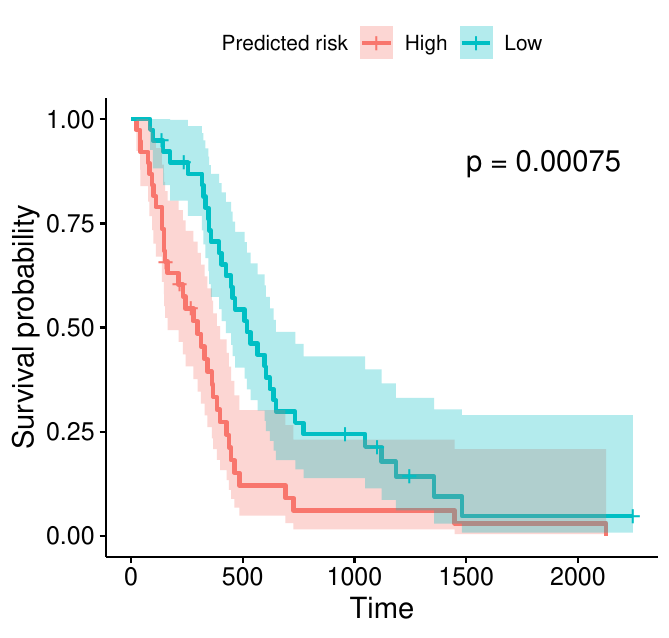}
            \caption{FCoxPH model}
        \end{subfigure}
	    \caption{The Kaplan–Meier plots for the high- and low-risk groups of the Cox proportional hazards (CoxPH) model and functional Cox proportional hazards models using SECT (FCoxPH-SECT) and persistent homology (FCoxPH) of glioblastoma multiforme patients. 
	    }
	    \label{fig:KMplot2}
	\end{figure}

Figure~\ref{fig:KMplot2} shows the Kaplan-Meier plots of the high- and low-risk groups of the three models. The p-values of the log-rank tests are $1\times10^{-2}$, $6\times 10^{-3}$ and $6\times 10^{-4}$ for the CoxPH, FCoxPH-SECT, and FCoxPH models, respectively. Both FCoxPH-SECT and FCoxPH models outperform the CoxPH model. Also, the FCoxPH model provides better separation than the FCoxPH-SECT model. The log-rank tests imply that the proposed persistent homology shape features provide additional information in predicting the overall survival prognosis of the GBM patients after taking into account the clinical predictors. The hazard ratios between the predicted high-risk and low-risk groups are $1.846$, $1.917$, and $2.176$ for the CoxPH, FCoxPH-SECT, and FCoxPH models, respectively. 

We note that the simulation study under false shape information similar to Section~\ref{subsec:falsepositive} is not conducted for the GBM application. The pixel-rearranged images have tumor pixels in an empty region because the 2D MRI images do not have information on the edges of the brain.

	\section{Conclusion}
	\label{sec:conclusion}
	
	In this article, we propose a new summary statistic that uses topological features to represent tumor shape patterns in medical images. We develop the distance transform for three-class pathology and two-class radiology images that reveal various tumor patterns. Persistent homology is applied to quantify tumor shape patterns, and the topological persistence information is represented as a function. The FCoxPH model is used to predict survival outcomes and enables potential clinical interpretation of topological shape features. The proposed topological shape features summarize tumor aggressiveness and improve prediction accuracy in the applications to lung adenocarcinoma and GBM images. The simulation studies suggest that the proposed approach is robust to false shape information.

	Our study leads to several future research topics.
	First, while we found a relationship between the topological shape features and survival prognosis, a complex relationship between the shape features, clinical variables, and genetic features is largely unknown.
	Similar to recent studies that investigate the relationships between imaging features and gene expressions \citep{li2019,moon2015}, one can explore the association with the topological shape features. 
	Second, one of the drawbacks of the proposed persistent homology approach is that it loses tumor location information. 
	This is suitable for the lung adenocarcinoma application because the images do not have a pre-determined direction. 
	However, the tumor spatial information could be important in some cancer image analyses, such as brain tumor \citep{bondy2008}.
	In the future, it would be useful to pair spatial information with the topological features computed by persistent homology.
	Lastly, our method can easily be applied to 3D medical images. In 3D image applications, zero-, one-, and two-dimensional topological shape features will be obtained, and their size is measured by the radius of the largest sphere.
	
	\section*{Data and Code}
	R and Python code to reproduce results can be downloaded from the public repository: \url{https://github.com/chulmoon/TopologicalTumorShape}.

	\begin{supplement}
    \stitle{Additional Details and Figures} \sdescription{This file contains supplementary sections and figures.}
    \end{supplement}

	\bibliographystyle{imsart-nameyear}
	\bibliography{References}

\end{document}